\documentclass{elsarticle}

\usepackage{times}
\usepackage{amsfonts}
\usepackage{hyperref}
\usepackage{latexsym}
\usepackage{todonotes}
\usepackage{multirow}
\usepackage{amsmath}
\usepackage{pgf}
\usepackage{tikz}
\usepackage{dirtytalk}
\usepackage{quoting}
\usepackage{pgfplots}
\usepackage[named]{algo}
\usepackage{enumitem}

\journal{Journal of Cognitive Systems Research}

\begin{document}

\begin{frontmatter}

\title{A Study of Metrics of Distance and Correlation Between Ranked Lists for Compositionality Detection}

\author{Christina Lioma and Niels Dalum Hansen}
\address{Department of Computer Science, University of Copenhagen, Denmark}
\ead{\{c.lioma,nhansen\}@di.ku.dk}

%
%

\begin{abstract}
Compositionality in language refers to how much the meaning of some phrase can be decomposed into the meaning of its constituents and the way these constituents are combined.
Based on the premise that substitution by synonyms is meaning-preserving, compositionality can be approximated as the semantic similarity between a phrase and a version of that phrase where words have been replaced by their synonyms. Different ways of representing such phrases exist (e.g., vectors \cite{kiela-clark:2013:EMNLP} or language models \cite{LiomaSLH15}), and the choice of representation affects the measurement of semantic similarity. 

We propose a new compositionality detection method that represents phrases as ranked lists of  term weights. Our method approximates the semantic similarity between two ranked list representations using a range of well-known distance and correlation metrics. In contrast to most state-of-the-art approaches in compositionality detection, our method is completely unsupervised. Experiments with a publicly available dataset of 1048 human-annotated phrases shows that, compared to strong supervised baselines, our approach provides superior measurement of compositionality using any of the distance and correlation metrics considered.  
\end{abstract}


\end{frontmatter}

\section{Introduction}
Compositionality in natural language describes the extent to which the meaning of a phrase can be decomposed into the meaning of its constituents and the way these constituents are combined.  Automatic compositionality detection has long received attention (see, for instance, \cite{kiela-clark:2013:EMNLP}, \cite{lin:1999:ACL}, 
\cite{baldwin-EtAl:2003:MWE}). One popular compositionality detection technique is to replace the constituents of a phrase by their synonyms (one at a time), obtaining a new ``perturbed'' phrase, and to measure the semantic distance of the resulting phrase to the original phrase: the smaller the semantic distance, the higher the degree of compositionality in the original phrase. Table \ref{tab:example} illustrates this point.

\begin{table}
\centering
\small
\resizebox{120mm}{!}{
\begin{tabular}{| c | c | c | c |}
\hline
Original Phrases & Perturbed Phrases & Semantic Distance & Compositionality\\
\hline
\texttt{red car}	&\texttt{scarlet car, red vehicle}	&small	&high\\
\texttt{red tape}	&\texttt{scarlet tape, red ribbon}	&large	&low\\
\hline
\end{tabular} 
}
\caption{Examples of a very compositional (\texttt{red car}) and a non-compositional (\texttt{red tape}) phrase, and their perturbations.}
\label{tab:example}
\end{table}

Different variants of this substitution-based approach to compositionality detection exist (overviewed in Section \ref{s:rw}), which differ in their measurement of semantic similarity between the original and the perturbed phrase. One approach is to represent the original phrase and its perturbed phrases as vectors, whose elements are contextual terms (i.e. frequently co-occurring terms) extracted from their respective distributional semantics in some appropriate corpus. Then, vector distance can be measured using, for instance, the cosine of the  angle of the vectors. Another approach is to represent the original and perturbed phrases as language models and approximate their similarity, e.g., using Kullback-Leibler divergence. These are mostly \emph{supervised} approaches that consider the aggregated distance or divergence (over all the elements of the vectors or language models) as an (inverse) approximation of semantic similarity. 

We conjecture that it is not necessary to use all the elements of such vectors or language models to compute semantic similarity, but that it suffices to use solely the most semantically salient elements. If this conjecture is true, the potential benefit is that semantic similarity (and any derived analysis thereof) can be done more efficiently as, e.g., the involved computations will be done on lower dimension elements. Motivated by this, we propose the following \emph{alternative approach to compositionality detection}: we represent the original phrase and its perturbed phrases as ranked lists, containing only the term weight (e.g., TF-IDF \cite{ksj72}) of their contextual terms. We rank the elements of each list decreasingly by this term weight. This allows to compute the semantic similarity between the phrase and its perturbed phrases using a range of different metrics designed for ranked list comparison. Different such metrics emphasise different aspects of the ranked lists, such as the weight of the elements in the list, or their position in the ranking, or the similarity between elements. 
As metrics of distance and correlation between ranked lists tend to be parameter-free, our approach is completely unsupervised.

{\bf Our contribution:} We propose a novel formulation of compositionality detection as ranked list similarity that is completely unsupervised. We perform an empirical study of how this formulation fares when a number of different distance and correlation metrics for ranked lists are used. The results of the study show that our method performs better than strong, recent, supervised baselines.

\section{Related Work on Compositionality Detection}
\label{s:rw}

We divide related work on compositionality detection into two broad categories: 
\begin{itemize} 
\item approaches estimating the similarity between a phrase and its components (mostly earlier), and
\item approaches estimating the similarity between a phrase and a perturbed version of that phrase where terms have been replaced by their synonyms (more recent).
\end{itemize}

Baldwin et al. \cite{baldwin-EtAl:2003:MWE} use Latent Semantic Analysis (LSA) to calculate the similarity between a phrase and its components, reasoning that higher similarity indicates higher compositionality. They extract contextual vectors for the terms in the phrase, and represent them as points in vector space. They measure the similarity between two vectors as the cosine of the angle between them. Katz and Giesbrecht \cite{katz-giesbrecht:2006:MWE} present a similar idea (with a slightly different choice of methods and with German data). Venkatapathy and Joshi \cite{venkatapathy-joshi:2005:HLTEMNLP} also present a similar idea, by extending the LSA context vectors of Baldwin et al. \cite{baldwin-EtAl:2003:MWE} with collocation features (e.g. phrase frequency, point-wise mutual information) extracted from the British National Corpus. More recently, Reddy et al. \cite{reddy-mccarthy-manandhar:2011:IJCNLP-2011} define a term literality score as the similarity between a phrase and its constituent contextual vectors. They use different vectorial operations to estimate the semantic distance between a phrase and its individual components, from which they deduce compositionality.

Closer to ours is the work of Kiela and Clark \cite{kiela-clark:2013:EMNLP}, who detect non-compositionality based on the earlier hypothesis that the mutual information between the constituents of a non-compositional phrase is significantly different from that of a phrase created by substituting terms in the original phrase by their synonyms \cite{lin:1999:ACL}. 
They represent phrases by their context vectors. Using standard vectorial similarity, this model slightly outperforms that of McCarthy et al. \cite{mccarthy-venkatapathy-joshi:2007:EMNLP-CoNLL2007} and Venkatapathy and Joshi \cite{venkatapathy-joshi:2005:HLTEMNLP}. A recent variation of this idea replaces the context vectors with language models and computes their Kullback-Leibler divergence to approximate their semantic distance \cite{LiomaSLH15}. However, the accuracy of this approach has not been evaluated. 

Further approaches to compositionality detection also exist. For instance, Cook et al. \cite{cook-fazly-stevenson:2007:ACL07-MWE} use syntax to identify non-compositionality in verb-noun phrases. They reason that compositional expressions are less syntactically restricted than non-compositional ones, as the latter tend to occur in a small number of fixed syntactic patterns. 
Along a similar line, McCarthy et al. \cite{mccarthy-venkatapathy-joshi:2007:EMNLP-CoNLL2007} consider the selectional preferences of verb-object phrases. 

Lastly, compositionality detection has also been studied using representation learning of word embeddings. 
Socher et al. \cite{socher-EtAl:2012:EMNLP-CoNLL} 
present a recursive neural network (RNN) model that learns compositional vector representations for phrases and sentences of arbitrary syntactic type and length. They use a parse tree structure and assign a vector and a matrix to every node: the vector captures the meaning of the constituent, while the matrix captures how it changes the meaning of neighboring words or phrases. Mikolov et al. \cite{MikolovSCCD13} extend a word-based skip-gram model that learns non-compositional phrases by being trained on phrases (treated as individual tokens) as opposed to individual words. The training phrases are extracted from a corpus using a threshold on the ratio of their bigram over (product of) unigram counts. Along a similar line, Salehi et al. \cite{salehi-cook-baldwin:2015:NAACL-HLT} use the word embeddings of Mikolov et al. \cite{DBLP:journals/corr/abs-1301-3781} with several vectorial composition functions to detect non-compositionality. Yazdani et al. \cite{yazdani-farahmand-henderson:2015:EMNLP} also learn semantic composition and detect non-compositional phrases as those that stand out as outliers in the process. They examine various composition functions of different levels of complexity and conclude that complex functions such as polynomial projection and neural networks can model semantic composition more effectively than the commonly used additive and multiplicative functions.

To our knowledge, no prior work on compositionality detection has used ranked list distance or correlation.

\section{Ranked lists for compositionality detection}
\label{s:model}

\subsection{Problem formulation}
The starting point of our method is the substitution-based approach to non-composi\-tionality detection of Kiela and Clark \cite{kiela-clark:2013:EMNLP}. Given a phrase, each term is replaced by a synonym (one at a time) producing perturbations of the original phrase by substituting synonyms. The semantic similarity between the original phrase and its perturbations is then assumed to be proportional to the degree of compositionality of the original phrase. This idea is in fact a modern implementation of Leibniz's principle of intersubstitutivity (\textit{salva veritate}) to detect irregular  composition of meaning, which posits that terms which can be substituted for one another without altering the truth of any statement are the same (\textit{eadem}) or coincident (\textit{coincidentia}).

\subsection{Our compositionality detection method}
\label{alg}
Kiela and Clark \cite{kiela-clark:2013:EMNLP} represent the original phrase and its perturbations as vectors (of their contextual semantics) and compute their semantic distance using cosine similarity, while Lioma et al. \cite{LiomaSLH15} represent them as language models and approximate their semantic distance using Kullback-Leibler divergence\footnote{Strictly speaking, Kullback-Leibler divergence is not a distance.}. Instead, we represent the original phrase and its perturbations as \textit{ranked lists}, and use their (inverse) distance or correlation to measure compositionality. The elements of each ranked list are term weights (e.g. TF-IDF) of their contextual terms that are typically represented in vectors. 
The ranking in each list is by descending term weight, and hence more informative terms are represented earlier in the list.

The high-level steps of our method are displayed in Algorithm I (Table \ref{alg1}).
\begin{table}[!htbp]
\begin{algorithm}{RankListComp}{
\qprocedure{RankListComp}
\qinput{Phrase $p$}
\qinput{Corpus $C$}
\qoutput{Compositionality score $comp(p)$}
}
Find synonym $\widehat{t}$ of each term $t \in p$\\
\qfor each $t $ and $\widehat{t}$\\
	$\mathcal{C}$ $\leftarrow$ get context terms from $C$\\
	\qfor each context term $i \in \mathcal{C}$\\		
		compute term weight $g_i$ \qendfor	
	\qendfor\\
\qfor each $\widehat{t}$\\
	Perturbed phrase $\widehat{p} \ni  \{\widehat{t},\left\vert{p}\right\vert-1$ original terms $t_o$\}\qendfor\\

List $L_p \leftarrow$ sort \{$g(i) \forall i \in \mathcal{C}^{\forall t \in p}$\}\\
List $L_{\widehat{p}}  \leftarrow$ sort \{$g(i) \forall i \in \mathcal{C}^{\widehat{t} \land \forall t_o}$\}\\
\qreturn $\frac{1}{|L_{\widehat{p}}|} \sum^{|L_{\widehat{p}}|}$ similarity($L_p$, $L_{\widehat{p}}$)	
\end{algorithm}
\caption{Algorithm I.}
\label{alg1}
\end{table}

The ranked lists of the original and perturbed phrase appear in lines 8 -- 9 of the above algorithm, respectively. Central in the computation of compositionality is the choice of similarity metric (line 10). A number of different ways to compute this similarity (as distance or correlation) between two ranked lists exist in the literature (for the problem of comparing ranked lists in general, and not for our specific problem formulation). Some metrics consider the aggregate distance over the whole list, whereas other metrics emphasise specific list characteristics, such as the weight, position, or similarity between elements. 
Next, we present an overview of these metrics. 

\subsection{Metrics of distance and correlation between ranked lists}
\label{s:metrics}

Let $m$ and $n$ be positive integers.
We consider lists $R_1 = [v_1,\ldots,v_m]$ and $R_2 = [w_1,\ldots,w_n]$ where 
$v_1,\ldots,v_m,w_1,\ldots,w_n \in \mathbb{R}_+ \cup \{0\}$ (i.e., each element in each list is a non-negative real number). 
We assume that $n \geq m$ (i.e., $R_2$ is possibly longer than $R_1$) and
$v_1 \geq v_2 \geq \cdots \geq v_m$ and $w_1 \geq w_2 \dots \geq w_n$ (i.e., both lists are ordered non-increasingly).

In our problem formulation, $R_1$ and $R_2$ represent \emph{ranked lists} of term weights. In principle, the lists may be of unequal length, depending on the implementation of extracting term weights and/or the corpus statistics from which we extract contextual terms. Note also that there is no a priori relationship between the elements of $R_1$ and $R_2$. Indeed in extreme cases we may
have $\{v_1,\ldots,v_m\} \cap \{w_1,\ldots,w_n\} = \emptyset$. 

Next, we review metrics on lists that have been, or could be, used to measure some notion
of ``(dis)similarity'' between ranked lists. We partition these metrics into three general classes:
\begin{itemize}

\item Class I: Metrics that can natively compute differences between ranked lists of \textbf{unequal length}. This is the scarcest
class, consisting primarily of modern metrics specifically devised to tackle ranked lists.

\item Class II: Metrics that can compute differences between ranked lists of \textbf{equal length}. Such metrics can be applied to ranked lists on unequal length only if the \emph{shortest} list ($R_1$ above) is padded with $n-m$ zeroes at the end, or if the \emph{longest} list ($R_2$ above) is pruned to the length of $R_1$. Both of these options can only work when they do not alter the semantics of the data represented in the list. Most of these Class II metrics are classic
metrics on the vector space $\mathbb{R}^n$. 

\item Class III: Metrics that can compute differences between ranked lists which constitute \textbf{permutations
of a set of $n$ elements}. Formally, let $[ n ] = \{ 1, ..., n\}$, and let $S_n$ be the set of permutations on $[n]$; for
two elements $\sigma, \pi \in S_n$, a metric $d$ assigns a non-negative real number $d(\sigma,\pi)$ as a ``distance'',
with $d(\sigma,\pi) = 0$ if{f} $\sigma = \pi$.


\end{itemize}

Most of the metrics we have come across are Class III. However, as the ranked lists $R_1$ and $R_2$ of term weights that we are interested in may contain different elements and
may be of different length, Class III metrics are not directly applicable to our setup of using ranked lists: if $R_1$ and $R_2$ were of equal length and contained the same elements, the fact that they are both ordered implies that they would correspond to the exact same permutation of $n$ elements, hence that their distance would be $0$ (because they would be identical). 
Even though Class III metrics cannot be used in our setup, for completeness we briefly outline them in Table \ref{tab:classIII}. Next we focus on Class I and II metrics of distance (Section \ref{ss:distance}) and correlation (Section \ref{ss:correlation}).


\begin{table*}
\centering
\small
\resizebox{120mm}{!}{
\begin{tabular}{| l |}
\hline
$\bullet$ \textbf{Spearman's footrule} (a.k.a.  $l_1$-distance): total element-wise displacement between \\
two ranked lists. Variations include: weighted, positional, element similarity, and \\
generalised (weighted + positional + element similarity) \cite{KumarV10}\\
$\bullet$ \textbf{Kendall's $\tau$}: total number of pairwise inversions between two ranked lists. Variations\\
include: weighted, positional, element similarity,  generalised (weighted + positional\\
+ element similarity) \cite{KumarV10}, version with penalty parameter for swaps early in the \\
permutation \cite{FaginKS03}, and weighted generalisation for ties \cite{Vigna15}\\
$\bullet$ \textbf{Cayley distance}: minimal number of transpositions (permutations that swap two \\
 adjacent elements) needed to transform element $\sigma$ to element $\pi$  \cite{DezaH98}\\
$\bullet$ \textbf{Lee distance}: variant of $d_{\textrm{rank}}$ (Eq. 1) where differences $\vert \sigma(i) - \pi(i) \vert$ larger than $\lceil n/2 \rceil$\\ 
are instead counted as $n - \vert \sigma(i) - \pi(i) \vert$  \cite{DezaH98}\\
$\bullet$ \textbf{Expected weighted Hoeffding distance}: can handle partial or missing rank \\
information \cite{SunLC10}\\
\hline
\end{tabular} 
}
\caption{Class III metrics (on the set of permutations). The notation of Section \ref{s:metrics} is used.}
\label{tab:classIII}
\end{table*}

%
We now give a brief overview of the pertinent metrics.
%
We divide the metrics into distances (which, strictly speaking, should satisfy the conditions of being non-negative \textit{and} symmetric \textit{and} the shortest possible path between two points) and correlations.

\subsubsection{Distance metrics}
\label{ss:distance}

\paragraph{3.3.1.1 Rank or $l_1$ distance (Class III), Minkowski distances (Class II)}
Ciobanu and Dinu \cite{CiobanuDinu13} introduce the \emph{rank distance} $d_{\textrm{rank}}$ (identical to the well-known
$l_1$ distance, a.k.a. Spearman's footrule). $d_{\textrm{rank}}$ is the sum of absolute rank differences in the lists $R_1$ and $R_2$:
\begin{equation}
\label{eq:l1}
d_{\textrm{rank}}(R_1,R_2) = \sum_{i \in [n]} \vert \sigma(i) - \pi(i) \vert
\end{equation}

While the version of Ciobanu and Dinu \cite{CiobanuDinu13} is Class III, the $l_1$ distance can also easily be viewed as being in Class II. Indeed any of the Minkowski distances $l_p$ for $p \geq 1$ induce a distance metric in the usual sense on ranked lists of equal length:
\begin{equation}
l_p(R_1,R_2) = \left( \sum_{i=1}^n \vert _i - w_i \vert^p \right)^{1/p}
\end{equation}
A similar metric is the \emph{Chebyshev distance}, below.

\paragraph{3.3.1.2 Chebyshev or $l_{\infty}$ distance (Class II)}
The \emph{Chebyshev} or $l_{\infty}$ distance is the maximal absolute difference in rank between two equal-sized ranked lists:
\begin{equation}
\label{eq:cheb}
d_{\infty}(R1,R2) = \max_{i \in [n]} \vert v_i - w_i \vert
\end{equation}

\paragraph{3.3.1.3 CosRank distance (Class III, Class II)}
Dinu and Ionescu \cite{DBLP:conf/synasc/DinuI12} introduce the cosine rank distance $d_{\textrm{CosRank}}$
as the usual cosine distance in an $n$-dimensional vector space of the vectors $(\sigma(1),\ldots,\sigma(n))$
and $(\pi(1),\ldots,\pi(n))$:
\begin{equation}
d_{\textrm{CosRank}}(R_1,R_2) = \frac{\sum_{i \in [n]} \sigma(i)\pi(i)}{\sum_{i\in[n] i^2}}
\end{equation}
The usual cosine similarity can also be used on the $n$-dimensional vector space $\mathbb{R}^n$,
hence on equal-sized ranked lists:
\begin{equation}
\label{eq:cosrank}
d_{\textrm{CosRank}}(R_1,R_2) = \frac{R_1 \cdot R_2}{\sqrt{R_1 \cdot R_1}\sqrt{R_2 \cdot R_2}}
\end{equation}
\noindent where $\cdot$ is the dot product of vectors.

\paragraph{3.3.1.4 Hamming distance (Class II)}

The \emph{Hamming distance} $d_H(R_1,R_2)$ between equal-sized ranked lists $R_1$ and $R_2$  is the number of indices $i$ where $v_i \neq w_i$; using
Kronecker's delta $\delta_{ij}$, we can write:
\begin{equation}
\label{eq:ham}
d_H(R_1,R_2) = \sum_{i=1}^n (1 - \delta_{v_iw_i})
\end{equation}
\noindent (equivalently, $d_H(R_1,R_2) = \vert \{i \in [n] : v_i \neq w_i\}\vert$).

%

%

\paragraph{3.3.1.5 Hausdorff distance (Class I)}
If a $d$ metric on elements is given, the Hausdorff distance is a particular metric computing distances between certain \emph{sets} of elements. For instance, if one considers the usual Euclidean distance $\vert \cdot \vert$ on real numbers, the Hausdorff distance can be used directly on ranked lists of possibly distinct lengths:

\begin{small}
\begin{multline}
\label{eq:haus}
d_{\textrm{Haus}}(R_1,R_2) = 
\max\left\{\max_{v_i \in R_1} \min_{w_j \in R_2} \vert v_i - w_j \vert , \max_{w_j \in R_2} \min_{v_i \in R_1} \vert v_i - w_j \vert \right\}
\end{multline}
\end{small}

The Hausdorff distance is thus the \emph{largest} difference in term weight from a term weight $v$ in $R_1$ to the term weight in $R_2$ that is closest to $v$.


%

\subsubsection{Correlation metrics}
\label{ss:correlation}
Ranked list similarity can also be estimated using \emph{correlation metrics} that generally do not always adhere to all of the classic
axioms required by distance metrics in mathematics. 


\paragraph{3.3.2.1 Pearson correlation coefficient (Class II)}
The Pearson (product-moment) correlation coefficient quantifies the linear dependence between two variables.
It is the ratio of the covariance of the variables to the product of their standard deviation, easily applicable
to two equal-sized ranked lists by using the formula for a sample:

{\footnotesize
\begin{multline}
\label{eq:pearson}
d_{\textrm{Pear}}(R_1,R_2) = 
\frac{n \sum_{i \in [n]} v_iw_i - \sum_{i \in [n]} v_i \sum_{i \in [n]} w_i}%
{\sqrt{n \sum_{i \in [n]} v_i^2 - \left( \sum_{i\in [n]} v_i \right)} \sqrt{n \sum_{i \in [n]} w_i^2 - \left( \sum_{i\in [n]} w_i \right)}} 
\end{multline}
}

Note that Etesami et al. \cite{EtesamiG16} introduce it as a ``Class III'' distance, i.e. to compare different rankings of the same group of items. Nevertheless, the usual Pearson correlation can be used directly, as it can be computed for samples, not just for random variables.

\paragraph{3.3.2.2 AP correlation coefficient (Class II)}
Yilmaz et al. \cite{YilmazAR08} present a correlation metric, the \emph{AP metric}, inspired by Kendall's $\tau$, that particularly penalizes differences in the top-ranked items. While the metric does not natively apply to lists containing distinct items, it is easy to adapt it to do so.

Let us define $\tau_{ap}(R_1 \vert R_2) = p' - (1-p')$
where 
\begin{equation}
\label{eq:ap}
p' = \frac{1}{n-1} \sum_{i=2}^n \frac{C(i)}{i-1}
\end{equation}
\noindent where $C(i)$ is the number of items in $R_1$ occurring earlier than $i$ whose value is greater than or equal
to the value of the $i$ item in $R_2$ (i.e., in our case $C(i) = \max_{j \in [n], j \leq i} v_j \geq w_i\}$).
$\tau_{ap}(R_2 \vert R_1)$ is defined symmetrically \textit{mutatis mutandis}.

The distance $\textrm{symm}\tau_{ap}(R_1,R_2)$ is now the following \emph{symmetric} function:
\begin{equation}
\textrm{symm}\tau_{ap}(R_1,R_2) = \frac{\tau_{ap}(R_1 \vert R_2) + \tau_{ap}(R_2\vert R_1)}{2}
\end{equation}

\paragraph{3.3.2.3 Further Class II metrics}
Two more correlation metrics that can compute differences between ranked lists of equal length (Class II metrics) are 
the \textbf{Hirschfeld-Gebelein-R\'{e}nyi maximal correlation} 
and the \textbf{Maximal Rank Correlation} \cite{EtesamiG16}. However, the Hirschfeld-Gebelein-R\'{e}nyi maximal correlation is a function that, by definition, requires a probability distribution. We do not have such a probability distribution in our setup. The reason we can do without a probability distribution for the Pearson correlation is that it is possible to compute covariance, standard deviation etc. for a sample --- but doing so for the Hirschfeld-Gebelein-R\'{e}nyi maximal correlation and the Maximal Rank Correlation does not seem to make sense without a probability distribution (in particular since the definitions quantify over all functions between two spaces). Also note that no polynomial-time algorithm is known for computing Maximal Rank Correlation, even if an efficiently computable probability mass function is given as input (for instance, by approximating the probability mass functions from relative frequencies).

%
%
For compositionality detection, we use all the above distance and correlation metrics (except those discussed in Section 3.2.2.3) to compute the (dis)similarity in line 10 of our algorithm, i.e. the semantic distance between an original phrase and its perturbed version.

Next, we describe in detail the implementation steps of our compositionality detection method.

\section{Implementation}
We refer the reader to the algorithm of our compositionality detection method (Algorithm I: RankListComp) displayed in Table \ref{alg1}. The remainder explains how each line in the algorithm is implemented. 

\subsection{Synonym extraction (Algorithm I, line 1)}
The input is some phrase, and the goal is to measure its compositionality. Given an input phrase, for each of its terms, we fetch from WordNet all its hypernyms, and for each hypernym we select all hyponyms. If there are no hypernyms or hyponyms, we fetch all synonyms. We retrieve from the TREC disks 4-5 corpus\footnote{The TREC disks 4-5 corpus is available from: \url{http://www.nist.gov/tac/data/data_desc.html} and contains text from a variety of sources: the 103$^{rd}$ Congressional Record, the 1994 Federal Register, the 1992-1994 Financial Times, the 1996 Foreign Broadcast Information Service, and the 1989-1990 Los Angeles Times. We use this corpus because of its coverage and diversity. Any other corpus of reasonable coverage can be used alternatively.} all documents that contain any of these hypernyms  and hyponyms (and synonyms, if applied), and select the 100 documents where the original term \textit{and} its hypernyms \textit{and} its hyponyms (or synonyms) occur most often. We then select the single hypernym or hyponym (or synonym) that occurs most often in most of these 100 documents. We consider this term as a ``near-synonym'' to the original input term.


\subsection{Contextual terms (Algorithm I, line 3)}
For each term (in the original phrase + ``near-synonym'') we identify a context window of $\pm 5$ terms around it from TREC disks 4-5 (i.e. the 5 terms preceding the term and 5 terms following it). We neither remove stopwords, nor apply stemming. 
The above extraction of context windows can give a different number of context terms for different words. 
Other sizes of context windows can also be used; we use $\pm 5$ terms, following prior work \cite{LiomaSLH15}.


\subsection{Term weights (Algorithm I, line 5)}
We compute the TF-IDF score\footnote{Any other reliable term weighting score can be used alternatively, for instance any of the simpler \cite{LiomaB09,LiomaO07,LiomaK:2008} or more elaborate \cite{BlancoL07,BlancoL12,LiomaLL12} formulations in the literature.} of each term $t$ in the context windows as:
\begin{equation}
\sum_{d \in n_t} \frac{f(t,d)}{|d|} \cdot log_e \frac{N}{n_t}
\end{equation}
\noindent where $f(t,d)$ is the frequency of term $t$ in document $d$, $|d|$ is the total number of terms in document $d$, $N$ is the total number of documents in the collection, and $n_t$ is the total number of documents that contain $t$ in the collection. We compute TF-IDF using the corpus statistics of TREC disks 4-5 (unlike  \cite{kiela-clark:2013:EMNLP}, who use an adapted version of TF-IDF on the statistics of the context windows). We do this because we aim to compute TF-IDF scores reflecting how generally informative a word is, hence the bigger the corpus used to estimate this, the more representative the resulting scores will be (as long as the corpus has good coverage and representativess).

%


\subsection{Ranked lists (Algorithm I, lines 8 -- 9)}
We create ranked lists of TF-IDF scores for each term in the original phrase and its ``near-synonyms''. We then combine the ranked lists of TF-IDF scores extracted for each term, into a single ranked list for the whole original phrase (and separately for each perturbed phrase), by simply appending their respective TF-IDF scores to one list and sorting them by their TF-IDF. We do not keep duplicate elements. At this point, we have a single ranked list of TF-IDF scores per phrase (either original or perturbed). 

The length of the ranked lists may be different from term to term and may differ across phrases. Indeed the minimum and maximum length of ranked lists we observe per phrase are min=200 (for the perturbed phrase \texttt{musculus contractions} generated in response to the original phrase \texttt{muscle contractions}), and max=31309 (for the perturbed phrase \texttt{know 1} generated in response to the original phrase \texttt{know one}). The mean list length is 9259.  This variation in the length of our ranked lists practically means that we cannot always use the majority of distance and correlation metrics presented in Section \ref{s:metrics}, because they require equal-sized lists. To address this, we impose a fixed list length of maximum 1000 elements, i.e. we keep, at most, the top-1000 ranked elements in each list. If, when comparing two lists, one contains $<$1000 elements, we prune both lists to the length of the shortest list. The choice of 1000 as maximum length is an arbitrary choice and is not indicative of any tuning. Fixing the length of our ranked lists to 1000 implies that we represent each phrase by the top 1000 most informative terms found in their context windows. 

We compute the distance and correlation between the ranked list of TF-IDF scores of the original phrase and its substitution phrase using all the Class I \& II metrics presented in Section \ref{s:metrics}. This outputs a single score per phrase, which we treat as an (inverse for distances or direct for correlations) approximation of the degree of compositionality of that phrase. 

\section{Experimental evaluation}
We evaluate our compositionality detection method on a recent dataset of 1048 2-term phrases (noun-noun) \cite{farahmand-smith-nivre:2015:MWE}. This is the largest compositionality-annotated dataset we could find. In this dataset, each phrase has four binary compositionality human expert assessments. We report Spearman's $\rho$ correlation between our method's decisions on compositionality and the average of the four human annotations of compositionality of that dataset (Table \ref{tab:results}). 

We report, as state of the art baseline performance, the additive and multiplicative models of Reddy et al. \cite{reddy-mccarthy-manandhar:2011:IJCNLP-2011} and the best performing deep learning method (sparse interaction) of Yazdani et al. \cite{yazdani-farahmand-henderson:2015:EMNLP}. We do not reimplement these methods; we only report their published scores on the same dataset as per Yazdani et al. \cite{yazdani-farahmand-henderson:2015:EMNLP}.

We see in Table \ref{tab:results} that our approach outperforms all baselines using any of the considered metrics. Among these distance and correlation metrics, the Chebyshev and Hausdorff distances score the lowest, but still outperform the baselines. The Chebyshev and Hausdorff distances emphasise the maximal absolute difference in rank and maximal difference in TF-IDF score respectively. It thus appears that this emphasis on maximal differences is not optimal for this setup. One reason could be that we have trimmed our ranked lists to the top 1000 highest TF-IDF scores, and that among those highest TF-IDF scores, maximal differences may not be as noticeable, as on a much bigger range of TF-IDF scores that represent most levels of term informativeness (as opposed to just the most informative terms). 
 
 \begin{table}
\centering
\small
\begin{tabular}{| l | l | l |}
\hline
\multicolumn{2}{|c|}{\textbf{Supervised (baselines)}}\\
ADD (Reddy et al., 2011)	&0.21	\\
MULT (Reddy et al., 2011)		&0.09	\\
Deep Learning (Yazdani et al., 2015)				&0.41\\
\hline
\multicolumn{2}{|c|}{\textbf{Unsupervised (ours)}}\\
Rank or $l_1$ distance (Equation \ref{eq:l1})				&0.59\\
Chebyshev or $l_{\infty}$ distance (Equation \ref{eq:cheb})	&0.50\\
CosRank distance (Equation \ref{eq:cosrank})				&0.60\\
Hamming distance (Equation \ref{eq:ham})				&0.55\\
Hausdorff distance (Equation \ref{eq:haus})				&0.50\\
Pearson correlation coefficient (Equation \ref{eq:pearson})		&0.62\\
AP correlation coefficient (Equation \ref{eq:ap})				&0.58\\
\hline
\end{tabular} 
\caption{Spearman's $\rho$ correlation between system decisions and human annotations of compositionality (the higher, the better).}
\label{tab:results}
\end{table}

The best performing metric is the standard Pearson's correlation, followed closely by the CosRank distance. Pearson's correlation is the ratio of the covariance of the TF-IDF scores in the two lists over the product of their standard deviation. On a higher level of abstraction, Pearson's correlation can be seen as analogous to the cosine distance in vector space: the cosine distance measures similarity in vector space, by considering only the non-zero dimensions of (very often) sparse vectors. It is possible to normalise the attribute vectors by subtracting the vector means, in which case one can compute the \textit{centered cosine similarity}, which is equivalent to Pearson's correlation. Our second best metric, CosRank, is also analogous to the cosine distance as discussed in Section \ref{s:metrics} and also by Dinu and Ionescu \cite{DBLP:conf/synasc/DinuI12}. The fact that our two best performing metrics can be seen as analogous to the cosine distance in vector space indicates that not only the choice of measurement, but also the choice of representation, can greatly affect performance.

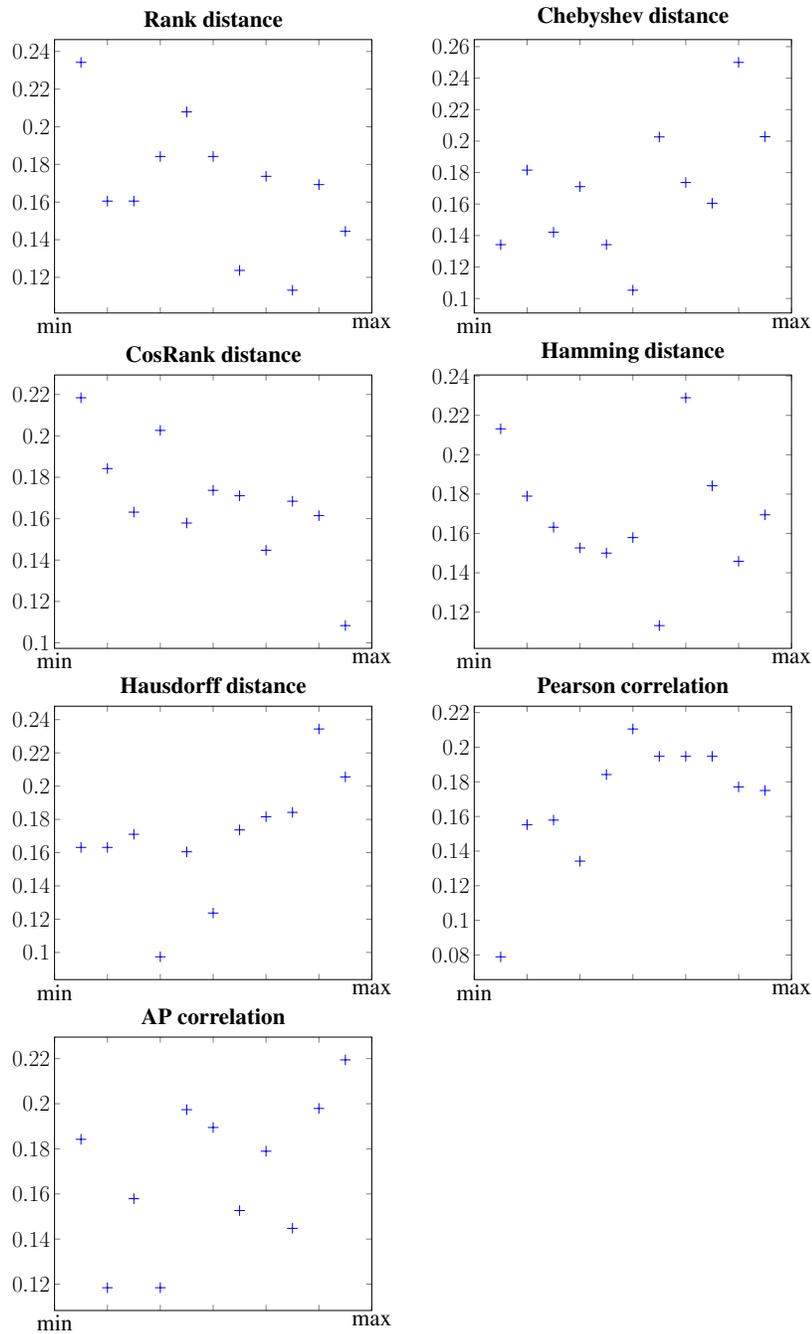
\begin{figure}
\begin{tabular}{cc}
\begin{tikzpicture}[baseline,scale=0.5]
\begin{axis}[
scale only axis,
xticklabels={,min, , , , , ,max},
font=\LARGE,
scaled y ticks = false,
y tick label style={/pgf/number format/fixed},
title=\textbf{Rank distance}
]
\pgfplotstableread{minBin.plot}\table 
\addplot+[mark=+,only marks, mark size=4.0, blue] table[x index=0,y index=1] from \table;
\end{axis}
\end{tikzpicture} 
&
\begin{tikzpicture}[baseline,scale=0.5]
\begin{axis}[
scale only axis,
xticklabels={,min, , , , , ,max},
font=\LARGE,
scaled y ticks = false,
y tick label style={/pgf/number format/fixed},
title=\textbf{Chebyshev distance}
]
\pgfplotstableread{cheBin.plot}\table 
\addplot+[mark=+,only marks, mark size=4.0, blue] table[x index=0,y index=1] from \table;
\end{axis}
\end{tikzpicture}  \\
\begin{tikzpicture}[baseline,scale=0.5]
\begin{axis}[
scale only axis,
xticklabels={,min, , , , , ,max},
font=\LARGE,
scaled y ticks = false,
y tick label style={/pgf/number format/fixed},
title=\textbf{CosRank distance}
]
\pgfplotstableread{cosrankBin.plot}\table 
\addplot+[mark=+,only marks, mark size=4.0, blue] table[x index=0,y index=1] from \table;
\end{axis}
\end{tikzpicture}  &
\begin{tikzpicture}[baseline,scale=0.5]
\begin{axis}[
scale only axis,
xticklabels={,min, , , , , ,max},
font=\LARGE,
scaled y ticks = false,
y tick label style={/pgf/number format/fixed},
title=\textbf{Hamming distance}
]
\pgfplotstableread{hamBin.plot}\table 
\addplot+[mark=+,only marks, mark size=4.0, blue] table[x index=0,y index=1] from \table;
\end{axis}
\end{tikzpicture}  \\
\begin{tikzpicture}[baseline,scale=0.5]
\begin{axis}[
scale only axis,
xticklabels={,min, , , , , ,max},
font=\LARGE,
scaled y ticks = false,
y tick label style={/pgf/number format/fixed},
title=\textbf{Hausdorff distance}
]
\pgfplotstableread{hausBin.plot}\table 
\addplot+[mark=+,only marks, mark size=4.0, blue] table[x index=0,y index=1] from \table;
\end{axis}
\end{tikzpicture}  &
\begin{tikzpicture}[baseline,scale=0.5]
\begin{axis}[
scale only axis,
xticklabels={,min, , , , , ,max},
font=\LARGE,
scaled y ticks = false,
y tick label style={/pgf/number format/fixed},
title=\textbf{Pearson correlation}
]
\pgfplotstableread{pearsonBin.plot}\table 
\addplot+[mark=+,only marks, mark size=4.0, blue] table[x index=0,y index=1] from \table;
\end{axis}
\end{tikzpicture}  \\
\begin{tikzpicture}[baseline,scale=0.5]
\begin{axis}[
scale only axis,
xticklabels={,min, , , , , ,max},
font=\LARGE,
scaled y ticks = false,
y tick label style={/pgf/number format/fixed},
title=\textbf{AP correlation}
]
\pgfplotstableread{apBin.plot}\table 
\addplot+[mark=+,only marks, mark size=4.0, blue] table[x index=0,y index=1] from \table;
\end{axis}
\end{tikzpicture}  &
\end{tabular}
\caption{\label{fig:bins} Non-compositionality score approximated by distance or correlation (x axis, binned) versus human compositionality score (y axis).}
\end{figure}

Figure \ref{fig:bins} plots the human compositionality annotations of the 1042 phrases (y axis, 0 = compositional, 1 = non-compositional) against the distance or correlation of our 7 metrics (x axis). The 1042 data points have been sorted per distance or correlation score and then binned into 11 bins. Each point in Figure  \ref{fig:bins} corresponds to the mean human compositionality score per bin. The number of bins has been decided using Scott's formula \cite{Scott79}: 
\begin{equation}
M = \frac{R}{3.49 s} N^{1/3}
\end{equation}

\noindent where $M$ is the number of bins, $R$ is the range, $N$ is the number of data points, and $s$ is the sample variance. This results in: 10 equal-sized bins of 95 phrases each, and 1 bin of the remaining 92 phrases (this is the bin with the highest distance or correlation).

In Figure \ref{fig:bins} ideally we would expect compositionality (x axis) to increase as distance decreases or as correlation increases. We see that this is indeed the case approximately for Rank CosRank, Hamming distance and for Pearson and AP correlation. However, for Chebyshev and Hausdorff, we see no such linear trend: the points are generally scattered, and seem to have more of an approximately ascending (as opposed to the expected descending) trend as the y axis increases. This finding agrees with the observation that these two metrics (Chebyshev and Hausdorff) performed the lowest among all our metrics in Table \ref{tab:results}. As discussed above, a reason why these two metrics underperform could be their emphasis on maximum score difference between the two lists (which is largely reduced when we trim lists to the top 1000).

\begin{table}
\centering
\small
\begin{tabular}{| l | l | }
\hline
\multicolumn{2}{|c|}{Non-compositional}\\
\hline
\texttt{freshman year} &highest Minkowski distance\\
\texttt{case law}  &highest Chebyshev distance\\
\texttt{body whorl} &highest CosRank distance\\
\texttt{umbrella organisation} &highest Hamming distance \\
\texttt{case law} &highest Hausdorff distance\\
\texttt{vice president} &lowest Pearson correlation \\
\texttt{chain smoker} &lowest AP correlation\\
\hline
\multicolumn{2}{|c|}{Compositional}\\
\hline
\texttt{mountain goat}  & lowest Minkowski distance\\
\texttt{picnic lunch} & lowest Chebyshev distance\\
\texttt{goose fossil} & lowest CosRank distance\\
\texttt{potato peeler} &lowest Hamming distance \\
\texttt{nightclub goer} &lowest Hausdorff distance\\
\texttt{school alumni} &highest Pearson correlation \\
\texttt{flight lessons} &highest AP correlation\\
\hline
\end{tabular} 
\caption{Examples from the top- and bottom-scored phrases by each of our metrics and by humans.}
\label{tab:examples}
\end{table}

Finally Table \ref{tab:examples} displays some examples of phrases that were annotated as most or least compositional \textit{both} by humans and also by our metrics (by having the highest/lowest distance/correlation respectively). We see that (a) compositional phrases tend to have more literal meanings that non-compositional, and (b) depending on the degree of (non-)compositionality, its detection may be a hard task even for humans (e.g. \texttt{case law, goose fossil}).

\section{Discussion of limitations}
\label{s:discussion}

As all substitution-based methods for compositionality detection, our method can also be criticised for being unable to discriminate non-compositional phrases from collocational phrases, because they both share the same property of non-substitutability (their constituents cannot be replaced with their synonyms) \cite{yazdani-farahmand-henderson:2015:EMNLP}. This criticism is related to the venerable \emph{principle of semantic substitutivity}, first formulated by Husserl (1913) \cite{sep-compositionality}: two phrases belong to the same semantic category if they are intersubstitutable within any meaningful expression \textit{salva significatione}. This principle is considered controversial, because there are many synonyms that are not everywhere intersubstitutable \cite{sep-compositionality}. To our knowledge, this remains an open problem in substitution-based compositionality detection.

Our method estimates the degree of compositionality of isolated phrases, following the current experimental practice of using mainly 2-term phrases. Applying the same method to phrases that are embedded into fully formed sentences may be problematic. For instance, problems may arise when the phrase whose compositionality we detect is a constituent of a larger expression. Borrowing an example from \cite{Geach65}, measuring the semantic similarity of:
\begin{quotation}
\texttt{Plato was bald}
\end{quotation}
with
\begin{quotation}
\texttt{baldness was an attribute of Plato}
\end{quotation}
could lead to misleading inferences about the semantic similarity of: 
\begin{quotation}
\texttt{the philosopher whose most eminent pupil was Plato was bald}
\end{quotation}
and 
\begin{quotation}
\texttt{the philosopher whose most eminent pupil was baldness was an attribute of Plato.}
\end{quotation} 
In this case, the second sentence, not only has a different meaning than the first sentence, but also is semantically non-sensical. It remains to be investigated to what extent and under which conditions phrase-level compositionality estimation can be applied to full sentences. The absence of large human-annotated bespoke datasets for this task remains a problem for such investigations.


\section{Conclusion}
We presented a method for estimating degrees of compositionality in phrases. Our method is based on the premise that substitution by synonyms is meaning-preserving \cite{lin:1999:ACL}, and estimates compositionality as the semantic similarity between a phrase and a version of that phrase where words have been replaced by their synonyms \cite{kiela-clark:2013:EMNLP}. Unlike previous approaches that represent such phrases (original and substitution-formed) as vectors or language models, we represent them as ranked lists. This ranked list representation is a novel contribution. The elements of these lists are contextual terms extracted from some appropriate corpus and ranked according to their TF-IDF (any other term informativeness score can be used). Moving to ranked list representations allows us to approximate the semantic similarity between two phrases using a range of well-known and, we argue, more refined distance and correlation metrics, designed specifically for lists. We review a number of these metrics and experimentally show that they outperform state-of-art baselines for this task.  

\section*{Acknowledgements} Work partially supported by C. Lioma's FREJA research excellence fellowship (grant no. 790095). We thank Jakob Grue Simonsen for thoughtful comments and valuable insights.

\section*{References}

\end{document}